\documentclass{acta_hungarica}

\actaVolume{19}
\actaNumber{11}
\actaYear{2022}

\begin{document}

\setcounter{page}{249}
\title{Current Safety Legislation of Food Processing Smart Robot Systems -- The Red Meat Sector}
\shorttitle{Current Safety Legislation of Food Processing Smart Robot Systems}
\author[1]{Kristóf Takács}
\author[2]{Alex Mason}
\author[3]{Luis Eduardo Cordova-Lopez}
\author[4]{Márta Alexy}
\author[5]{Péter Galambos}
\author[6]{Tamás Haidegger}

\affil[1]{Antal Bejczy Center of Intelligent Robotics, Óbuda University; kristof.takacs@irob.uni-obuda.hu}
\affil[2]{Norwegian University of Life Science (NMBU), As, Norway; alex.mason@nmbu.no and also with the Research and Development Department, Animalia AS, Oslo 0585, Norway}
\affil[3]{Norwegian University of Life Science (NMBU), As, Norway; luis.eduardo.cordova-lopez@nmbu.no }
\affil[4]{University Research and Innovation Center (EKIK), Óbuda University; alexy.marta@uni-obuda.hu }
\affil[5]{EKIK, Óbuda University and John von Neumann Faculty of Informatics, Óbuda University; peter.galambos@irob.uni-obuda.hu }
\affil[6]{EKIK, Óbuda University; tamas.haidegger@irob.uni-obuda.hu }

\shortauthor{K. Takács}

\maketitle
\begin{abstract}
Ensuring the safety of the equipment, its environment and most importantly, the operator during robot operations is of paramount importance. Robots and complex robotic systems are appearing in more and more industrial and professional service applications. However, while mechanical components and control systems are advancing rapidly, the legislation background and standards framework for such systems and machinery are lagging behind. As part of a fundamental research work targeting industrial robots and industry 4.0 solutions for completely automated slaughtering, it was revealed that there are no particular standards addressing robotics systems applied to the agri-food domain. More specifically, within the agri-food sector, the only standards existing for the meat industry and the red meat sector are hygienic standards related to machinery. None of the identified standards or regulations consider the safety of autonomous robot operations or human--robot collaborations in the abattoirs. The goal of this paper is to provide a general overview of the regulations and standards (and similar guiding documents) relevant for such applications, that could possibly be used as guidelines during the development of inherently safe robotic systems for abattoirs. Reviewing and summarizing the relevant standard and legislation landscape should also offer some instrumental help regarding the foreseen certification procedure of meat processing robots and robot cells for slaughterhouses in the near future.
\end{abstract} 


\section{Introduction}
In the EU (European Union), the CE mark (Conformité Européenne) is part of the EU’s harmonised legislation, signifying that a product has been assessed to comply with the high safety, health and environmental protection requirements set by the EU. CE mark must be obtained for every new electrical product sold in the EEA (European Economic Area), that supports fair competition too, since all manufacturers have to fulfill the requirements of the same set of rules~\cite{CEurl}. The approval procedure can be  managed by the manufacturer (this way, the CEO bears all legal responsibility), or by an independent certification body (which is called a Notified Body, in case registered in the EU). Upon the system assessment (carried out by either a Notified Body or the manufacturer), the main goal is to ensure the conformity of the product with the regulations (legal requirements) before the product is put on the European market.

By default, standards are voluntary for product manufacturers. They should be based on a consensus between industrial and academic experts, codifying already existing good practices, methods and general requirements. Nevertheless, by definition, they are intended to be the best available set of requirements toward a certain field; for instance, the safety aspects of a type of system, standards often serve as the basis of regulations enacted by lawmakers. For example, the ISO/IEC 60601-1 standard (International Organization for Standardization, International Electrotechnical Commission) regarding medical electrical equipment became the basis of the EC MD (European Commission Machinery Directive) and the subsequent MDD (Medical Device Directive). When a Notified Body is dealing with a new system, it is usually considering the recommendations of some non-compulsory standards during the system assessment as well. Therefore, developers and manufacturers should consider those non-compulsory standards from the early periods of development too. After all, better compliance may increase competitiveness~\cite{competitiveness2020}. Until today, the agri-food domain has not seen such structured, specific standards. The increasing autonomy of robots and robotic systems used in the industry has reasonably enhanced certification challenges, only recently evolved standards have the potential to address the safety concerns of this new approach -- in an application domain specific manner~\cite{chinzei2019}.

\begin{figure}[t]
    \centering
    \includegraphics[width=0.65\columnwidth]{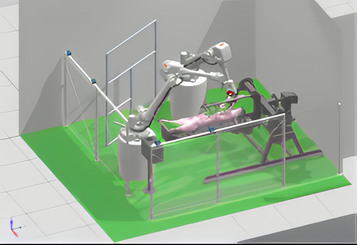}
    \caption{\centering Conceptual setup of the autonomous pig processing cell in the RoBUTCHER project. The carcass is handled by the actuated CHU (Carcass Handling Unit), the intelligent EOATs (knife, gripper) are fixed on the industrial ABB robots along with an RGB-D camera. \textit{Image credit: RoBUTCHER project, RobotNorge AS.}}
    \label{fig:MFC_RS}
\end{figure}

It is still an ongoing professional debate to unambiguously define what is a robot or a robotic system. However, efforts for standardization have been extensive in the robotics domain in the past three decades~\cite{Jacobs2018}. Traditionally, ISO standards have been providing guidance for safety of robots and robotic systems, and they have been forming the basis of the Machinery Directive~\cite{Directive2006}. The traditional ISO 8373 -- Robots and robotic devices – Vocabulary standard under ISO was published first in 1996, only referring to robots as \textit{Manipulating industrial robots}, but the document was later extended to all kinds of robots (in the ISO sense)~\cite{ISO8373_1996}. To incorporate all new domains, forms and applications of robots, TC299, the Technical Committee of ISO responsible for this topic, has revised the official definition of robots numerous times in the past decades. The key factors distinguishing robots from other machinery are autonomy, mobility and task-oriented behaviour. The current ISO definition of a robot as per ISO 8373 is~\cite{ISO8373}:
\begin{quote}
\textit{Programmed actuated mechanism with a degree of autonomy to perform locomotion, manipulation or positioning.}
\end{quote}
Wherein autonomy is defined as:
\begin{quote}
\textit{Ability to perform intended tasks based on current state and sensing, without human intervention.}
\end{quote}
Another modern holistic definition of a \textit{robot} from 2020 was given in the Encyclopedia of Robotics~\cite{Haidegger2021}:
\begin{quote}
\textit{A robot is a complex mechatronic system enabled with electronics, sensors, actuators and software, executing tasks with a certain degree of autonomy. It may be pre-programmed, teleoperated or carrying out computations to make decisions.}
\end{quote}
Robotics is advancing rapidly in practically all professional service domains, entering recently to the agri-food industry and within that, the meat sector as well~\cite{khamis2021robotics}. A prime example, the robot cell under development within the Ro-BUTCHER project (https://robutcher.eu) aims to carry out the primary cutting and manipulation tasks of a pig slaughtering~\cite{alvseike2020intact, CordisRobutcher}. The general purpose industrial robots in the cell are supported by RGB-D cameras, Artificial Intelligence, Virtual Reality, intelligent EOAT (End of Arm Tooling), telemanipulation and Digital Twin technology~\cite{de2020robotisation, Ian2022, Takacs2021Inner}. Accordingly, the designed system will be unprecedented in complexity and autonomy from the safety and legislation aspects. The robots in the Meat Factory Cell (MFC) will handle raw meat products intended for human consumption autonomously, however the risk of contamination due to the presence of the gastrointestinal tract is high. On the other hand, the applied EOAT (grippers, knives and saws) are designed for meat and bone cutting and gripping, making them highly dangerous for humans. Under the current  approach for classification within the related standards, the robot cell would be regarded as an industrial service robot application, meaning that it will still fall under the EC~MD (2006/42/EC) when assessing the safety assurance of the system (Fig.~\ref{fig:MFC_RS}).


This paper covers mainly the ISO standards, since they are commonly used in most sectors of the industry, practically world-wide accepted, and they have always been pioneers in the robotics field. Furthermore, ISO certification is often required by industrial customers due to its direct linkage to the EC~MD. It is worth mentioning however, that ISO does not act as a Notified Body, meaning that they do not issue certificates, only participate in the process by developing the international standards. The two main options for conformity assessment -- according to ISO -- are the followings:
\begin{itemize}
    \item \textbf{Certification} – the provision by an independent body in written assurance (a certificate) that the product, service or system in question meets specific requirements.
    \item \textbf{Accreditation} – the formal recognition by an independent body, generally known as an accreditation body that a certification body operates according to the international standards.
\end{itemize}

\newpage

 The clear, unambiguous and consistent use of the frequently occurring words, technical terms and expressions in the robot industry is essential, especially when documents with potential legal force are concerned. ISO 8373 states that: 

\begin{quote}\textit{This International Standard specifies vocabulary used in relation with robots and robotic devices operating in both industrial and non-industrial environment.}
\end{quote}
The standard was recently revised, the latest -- third -- version is ISO 8373:2021 that cancels and replaces the second edition originally from 2012)~\cite{ISO8373_2012,ISO8373}. Beside ISO standards, some important and relevant EU directives, guidelines and recommendations were also reviewed and will be summarized in this paper.

It is worth mentioning that a relevant Digital Innovation Hub (DIH), called agROBOfood was initiated in 2019, financed by the EU (https://agrobofood.eu/). Their motto being "Connecting robotic technologies with the agri-food sector", which means that they aim is to build a European ecosystem for adaptation of robotics and streamline standardization aspirations in Europe~\cite{agrobofood}. The agROBOfood consortium consists of 7 Regional Clusters involving 49 DIHs and 12 Competence Centers (as of late 2022), actively accelerating the agri-food sector’s digital transformation and robotization.

\section{Industrial robotics applied in the meat sector}

The automation in the meat industry has started long ago, however its pace is significantly slower compared to other industries. Machines dedicated to accomplish single tasks during the cutting processes were introduced in larger slaughterhouses relatively early, however only some simple, straightforward cuts could be automated by these machines. Several examples of such machinery were used and published world-wide in the late 20\textsuperscript{th} century, such as~\cite{longdell1994advanced} about the modern lamb and beef industry solutions at plants at New Zealand, or the world-wide review of machinery and level of automation of the meat industry by G.~Purnell~\cite{purnell1998robotic}.

Broader utilization of industrial robots and robot systems in the meat sector required the development of cognitive systems as well~\cite{echegaray2022meat}. Nowadays several manufacturers sell commercial intelligent systems for slaughterhouses (e.g., Frontmatec, Marel, Mayekawa, etc.) and the related research activity is significant as well. Mark Seaton from Scott Technology Limited published their experience gained in the last decades in \cite{seaton2022lessons}. Their key finding is that product consistency is the paramount advantage of the implementation of their robotised solutions, but shelf life, general product quality and workers' safety can improve as well. Their solutions include, e.g., X-ray based cutting prediction, de-boning with industrial robots and bladesaw that stops under a millisecond. Several review papers were published about the current stage and possibilities of meat industry automation e.g., the paper of Romanov et al.~about collaborative robot cells \cite{romanov2022towards}, or more general reviews by Khodabandehloo \cite{khodabandehloo2022achieving} or Esper et al.~\cite{de2020robotisation}.

\subsection{The RoBUTCHER project}

The RoBUTCHER research project is funded by the EU, and aims to develop the first entirely automated pig processing robot cell~\cite{mason2021thinking}. Therefore, it has been a priority for the study project to establish the framework of guiding documents~\cite{Ole2018}. The robot cell will carry out all primary steps of the pig-slaughtering with industrial robot arms (according to the EC MD), including the cutting of all four legs, the splitting of the carcass and the evisceration process~\cite{alvseike2020intact}. Beside the two robots, the cell consists of a motorized carcass handling unit (CHU), intelligent cutting and gripping tools, an RGB-D camera (fixed on one of the robots as well) and some other supplementary equipment (Fig.~\ref{fig:MFC_RS}). 

Modern image processing techniques are widely used in the meat sector to handle the natural variability of animals~\cite{szabo2022practical,alexy2022tracing}. The autonomous cutting begins with an imaging sequence, one of the robots moves the RGB-D camera (which is fixed on the shaft of the "smart knife"~\cite{mason2022smart}) to pre-defined positions around the carcass. In a simulation environment (powered by Ocellus: \textit{www.bytemotion.se}) a digital twin of the carcass is constructed from the images, where artificial intelligence (a set of image processing deep neural networks) calculates the desired gripping points and cutting trajectories on the carcass. Although the RoBUTCHER concept strictly rules out all kinds of collaborative behaviour between the machinery and humans, at this point, the supervising operator checks the predicted trajectories on the digital twin using Virtual Reality glasses~\cite{hansen2018collaborative}. They may accept the predicted cutting trajectories, request a new imaging and prediction, or may draw a 3D trajectory in the virtual environment that will be executed by the cutting robot. Thus -- in optimal case -- the cutting is performed completely without any interaction from human operators. Furthermore, even when the operator chooses to draw new trajectories there is no direct physical interaction, the robots and the operator never share the same workspace. In case of maintenance or any issue within the MFC that requires physical intervention, protective fencing with sensors will ensure that all machinery shuts down while anyone is working inside the cell.

This fundamental approach in abattoir-automation introduces several different challenges, not only from the engineering and developing aspects, but also from the safety and legislative side of the development. Automation and generally the robot industry are especially fast evolving and ever-changing fields, and since working together (in a collaborative way) at any level with robots and/or machines is always potentially dangerous, several directives and strict standards apply to robotized solutions \cite{romanov2022towards}.

\subsection{Guidance documents for safe industry applications}

Since the employment of general-purpose robot arms in cell-based automated raw-meat handling or animal slaughtering is unprecedented, no single standard exists that would regulate all aspects of this scenario. To have a thorough view on this domain's standardization, the complete list of Robotistry was scanned for relevant standards, along with traditional online search engines (e.g., Google Scholar, IEEExplore etc.)~\cite{robotistry}. 

\begin{figure}[t]
    \centering
    \includegraphics[width=\columnwidth]{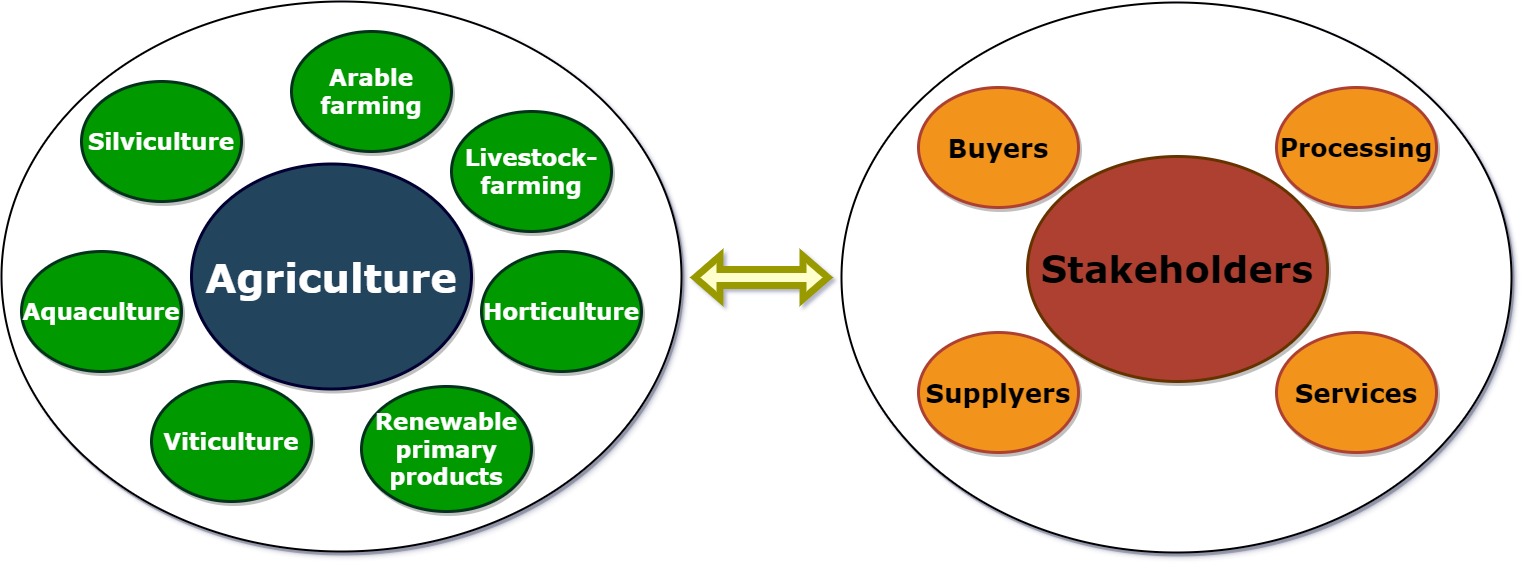}
    \caption{\centering Simplified structure of agricultural production categories and stakeholders according to the  Robotics 2020 MAR. Although the document mentions the meat sector within the agriculture section, it is not presented as a subcategory~\cite{Roadmap2020}.}
    \label{fig:agrisectors}
\end{figure}

An important document covering almost all aspects of robotics in the EU is the Robotics Multi-Annual Roadmap (MAR), which was also reviewed. However, automated slaughtering seems to be such a special part of robotics, that even the Robotics 2020 MAR does not cover it in detail~\cite{Roadmap2020}. The closest to slaughterhouse automation is the “Agriculture Domain” (Chapter 2.4), defined as:

\begin{quote}\textit{Agriculture is a general term for production of plants and animals by use of paramount natural resources (air, water, soil, minerals, organics, energy, information).}
\end{quote}
However, slaughtering itself does not appear in any of the subcategories (Fig.~\ref{fig:agrisectors}), animals are barely mentioned within the document. The only appearance of the meat sector is in the \textit{Food} section under the \textit{Manufacturing} sub-domains, where automation and machinery used for deboning and raw-meat handling are mentioned. This lack of recognition of the sector makes solution developers' task especially difficult.

In spite of not addressing the meat sector in depth, the recommendations in the “Safety design and certification” section are worth considering. Most of the statements and suggestions can be interpreted to the red meat domain too, although animal rights and animal welfare should always be kept in mind. The \textit{Hardware in Loop} and the \textit{Semantic Environment Awareness} sections contain interesting farming-related recommendations as well. The practical and beneficial application of simulations, planning systems, virtual models and semantic environment-representations are discussed, many of which are used in meat sector automation as well (and being used in the RoBUTCHER project).

The MAR document, however, only provides general guidelines, suggestions and best practises, while the certification is most crucial in the food industry. Therefore, in this section, the relevant robot industry-related standards will be discussed. 

ISO/IEC started to work on the integration of the new robotic application domains (e.g., collaborative robotics, medical robotics, self-driving cars) more than a decade ago. Numerous working groups (WGs) are active within the ISO/TC~299 Robotics Technical Committee, dealing with specific fields of robotics; e.g., \textit{Electrical interfaces for industrial robot end-effectors} (WG 9), \textit{Modularity for service robots} (WG 6) or \textit{Validation methods for collaborative applications} (WG 8). One of the most important and fundamental standards for practically every modern industrial automation project is the ISO 12100 Safety of machinery – General principles for design – Risk assessment and risk reduction standard, its latest version is the ISO 12100:2010~\cite{ISO12100}. The primary purpose of this standard is to provide engineers and system developers with an overall framework. The document acts as a guidance for decisions during the development of machinery, helping developers to design machines and whole systems that work safely while fulfilling their intended tasks. In spite of all this, at the beginning of the standard in the \textit{Scope} section it is highlighted that:

\begin{quote}\textit{It does not deal with risk and/or damage to domestic animals, property or the environment.}
\end{quote}
It is, therefore, clear that this comprehensive standard does not encompass specific information or advice tailored for meat sector automation projects.

ISO 12100:2010 offers a classification of the related safety standards that helps the identification of more specific, relevant ISO standards. This classification of standard-types is used in this paper as well~\cite{ISO12100}:

\newpage
\begin{itemize}
    \item \textbf{Type-A standards} (basic safety standards) giving basic concepts, principles for the design and general aspects that can be applied to machinery;
    \item \textbf{Type-B standards} (generic safety standards) dealing with one safety aspect or one type of safeguard that can be used across a wide range of machinery:
    \begin{itemize}
        \item \textbf{Type-B1 standards} on particular safety aspects (e.g., safety distances, surface temperature, noise);
        \item \textbf{Type-B2 standards} on safeguards (e.g., two-hand controls, interlocking devices, pressure-sensitive devices, guards);
    \end{itemize}
    \item \textbf{Type-C standards} (machine safety standards) dealing with detailed safety requirements for a particular machine or group of machines.
\end{itemize}

\noindent In this sense, ISO 12100 is not specifically robotics-standard, rather a more comprehensive one (a type-A standard), covering a wide range of machinery design safety – including robotic applications too. It is, however, intended to be used as the basis for preparation of type-B and type-C safety standards as well, that should be more specific to a given application. 

Regarding the given domain, presumably the most relevant type-B (more precisely type-B1) standard is the ISO11161:2007 Safety of machinery – Integrated manufacturing systems – Basic requirements~\cite{ISO11161}. As explained in its introduction, Integrated Manufacturing Systems (IMS) are very different in size, complexity, components, and they might incorporate different technologies that require diverse or specific expertise and knowledge, thus usually more specific (type-C) standards should be identified as well for a given application. As a consequence, this standard mainly describes how to apply the requirements of ISO 12100-1:2003, ISO 12100-2:2003 (and ISO 14121 Safety of machinery, which is currently inactive, due to its integration into ISO 12100) in specific contexts.

\begin{figure}[t]
    \centering
    \includegraphics[width=0.7\columnwidth]{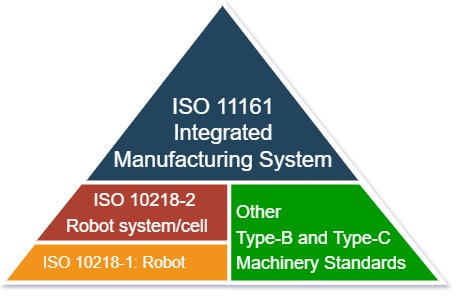}
    \caption{\centering Graphical representation of hierarchy between standards related to robot system/cell. ISO 11161 as a Type-A standard is on the top level relying on several different Type-B and Type-C standards.}
    \label{fig:relations}
\end{figure}

The ISO10218:2011 Robots and robotic devices – Safety requirements for industrial robots is a type-C standard, meaning that this document contains specific requirements and guidelines for system safety design that can potentially be used in meat sector automation projects. ISO10218:2011 consists of two main parts:
\begin{enumerate}
    \item Part 1: Robots~\cite{ISO10218_1};
    \item Part 2: Robot systems and integration~\cite{ISO10218_2}.
\end{enumerate}
While Part 1 only refers to the application of a single robot, Part 2 includes the peripheral elements connected to or working together with the robot(s) too. Part 2, in this sense, is typically more suitable for food-industry projects, since handling of carcasses and meat products usually requires complex EOAT and other external devices (e.g., sensors). Having more robots working together and/or employing external devices results in a ``robot system", thus the robot system specific problems (e.g., electrical connection between devices, overlapping workspaces, etc.) shall be considered  too~\cite{workspace2020, ross2021review}. Nevertheless, Part 2 of ISO10218:2011 naturally relies on information presented in Part 1, thus that should also be taken into consideration in all cases when Part 2 is being used. The relationship between the aforementioned ISO standards is shown in Fig.~\ref{fig:relations}.

\begin{figure}[t]
    \centering
    \includegraphics[width=\columnwidth]{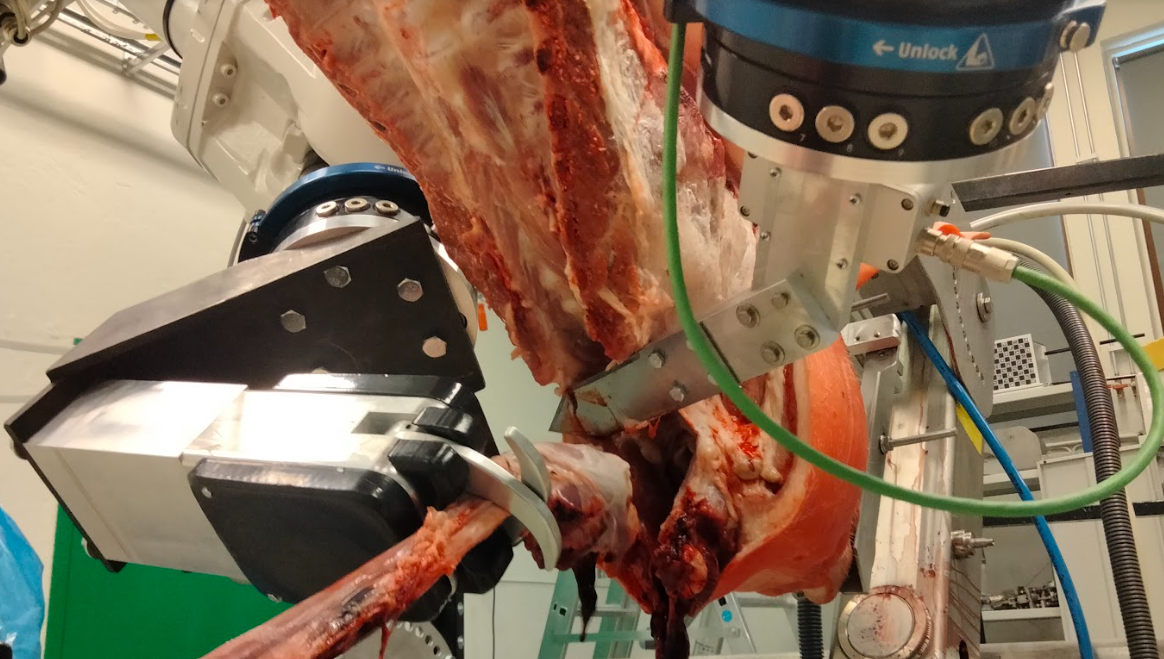}
    \caption{\centering  Typical intelligent EOAT for meat industry automation. The reinforced gripper with pointed claws and the sharp knife are "by design" dangerous robotic tools, even when the robot is not moving. \textit{Image credit: RoBUTCHER Project, Óbuda University \& NMBU} }
    \label{fig:tools}
\end{figure}

Another type-C standard, ISO/TR 20218-1:2018 Robotics – Safety design for industrial robot systems – Part 1: End-effectors should also be relevant. Meat-industry automation projects typically mean automated deboning and/or meat-cutting, both requiring sharp knives, saws and strong grippers, ``potentially dangerous end-effectors" in the wording of the standard (Fig.~\ref{fig:tools}.). Part 2 of ISO/TR 20218-1:2018 is about manual load/unload stations. This standard offers suggestions for applications where hazard zones are established around the robot(s). In such cases, access restriction to hazard zones and ergonomically suitable work spaces might be important, however, this is not the case in the described autonomous scenario~\cite{ISO20218}.

ISO/TR 20218-1:2018 Part 2 covers collaborative applications too, where human operators and robot systems share the same workspace. However, the recent increase of importance of collaborative robotics resulted in standalone standards for this new special field of robotics, the most significant ISO documents are ISO/TR 15066:2015 Robots and robotic devices – Collaborative robots and ISO/TR 9241-810:2020 Ergonomics of human--system interaction~\cite{ISO15066, ISO9241}. Beside these specific standards there are several projects and activities offering new solutions and assistance for collaborative robot system development, such as the EU funded \textit{COVR} project (https://safearoundrobots.com)~\cite{covr}.

Nevertheless, meat processing generally requires high payload robots, strong automated tools and single purpose machines that are intended to process and cut human-like tissues. The basic purpose of these devices self-evidently mean unacceptably high risk for any operator within the reach of the robots and the tools (the general workspace), regardless of how strict the safety regulations in place. Therefore, this paper (and the RoBUTCHER project) focuses on the completely automated slaughtering, telepresence of operators and strict physical perimeter guarding, excluding any type of collaborative work.

\subsection{ISO 10218: Robots and robotic devices — Safety \\requirements for industrial robots}

ISO 10218:2011 is arguably still the most important ISO standard in relevance of abattoir automation~\cite{ISO10218_1, ISO10218_2}. The latest version of the standard was published in 2011 (more than 10 years ago), but a new version (with a new title: Robotics — Safety requirements) is currently under development and should be published soon. ISO 10218:2011 mainly offers guidelines and requirements for inherent safe design of machinery (focusing on robots), presenting protective measures, foreseeable hazards and suggestions to eliminate, or at least reduce the risks associated with them. In the standards' terms \textit{hazards} are possible sources of harm, while the term \textit{risk} refers to hazard exposure.

ISO 10218:2011 Part 1: Industrial robots focuses on individual industrial robots, while Part 2: Robot systems and integration discusses the safety of robot systems and their integration into a larger manufacturing system. The most crucial statement in the standard is that any robot application (or robot system) should be designed in accordance with the principles of ISO 12100 for relevant and predictable hazards. It is worth mentioning as well that the standard emphasizes the fact that -- beside the several common and typical hazardous scenarios mentioned in the document -- task-specific sources of additional risk are present in most applications that should be examined in detail by the developers as well.

ISO 10218 includes important annexes. Annex A presents a list of common significant hazards, classified by their types (mechanical, electrical, etc.). For better understanding examples and potential consequences are presented along with the relevant clause in the standard for each hazard. According to the standard, a suitable hazard identification process should include risk assessment on all identified hazards. Particular consideration shall be given during the risk assessment to the followings:
\begin{itemize}
    \item The intended operations of the robot, including teaching, maintenance, setting and cleaning;
    \item Unexpected start-up;
    \item Access by personnel from any directions;
    \item Reasonably foreseeable misuse of the robot;
    \item The effect of failure in the control system;
    \item Where necessary, the hazards associated with the specific robot application.
\end{itemize}

\noindent Risks shall be eliminated, or at least reduced mainly by substitution or by design. If these preferred methods are not feasible, then safeguarding or other complementary methods shall be employed. Any residual risks shall then be reduced by other measures (e.g., warnings, signs, training).

ISO 10218 also suggests solutions in many relevant topics, such as:
\begin{itemize}
    \item robot stopping functions;
    \item power loss;
    \item actuating controls;
    \item singularity protection;
    \item axis limiting;
    \item safety-related control system performance.
\end{itemize}
Furthermore, the standard has a dedicated chapter (Information for use) to help preparing a useful and comprehensive documentation (called instruction handbook), using the suggested standard expressions, markings, symbols, etc. 

Further lists, specific instructions and specific detailed descriptions can be found in the appendix of ISO 10218:
\begin{itemize}
    \item Annex A: List of significant hazards;
    \item Annex B: Stopping time and distance metric;
    \item Annex C: Functional characteristics of three-position enabling device;
    \item Annex D: Optional features;
    \item Annex E: Labelling;
    \item Annex F: Means of verification of the safety requirements and measures.
\end{itemize}

\noindent Part 2 (Robot systems and integration) of ISO 10218 states that:
\begin{quote}
\textit{The design of the robot system and cell layout is a key process in the elimination of hazards and reduction of risks~\cite{ISO10218_2}.}
\end{quote}
Correspondingly, this part offers fundamental robot cell layout design principles, referring to different types of workspaces, physical limitations, perimeter safeguarding, manual intervention, human interfacing, ergonomics, etc. 

Different components of typical robot systems might fall under the scope of other standards too, thus Part 2 of ISO 10218 provides a useful list of those:
\begin{itemize}
    \item Equipotential bonding/earthing requirements (grounding): IEC 60204-1;
    \item Electric power: IEC 60204-1;
    \item Hydraulic power: ISO 4413;
    \item Pneumatic power: ISO 4414;
    \item Actuating control: IEC 60204-1;
    \item Emergency stop function: IEC 60204-1, ISO 13850, IEC 61800-5-2;
    \item Enabling devices: ISO 10218-1-Annex D.
\end{itemize}

\subsection{ISO/TR 20218-1:2018 Robotics — Safety design for \\industrial robot systems — Part 1: End-effectors}

ISO 20218-1:2018 is a relatively new TR (Technical Report) providing guidance for safe design and integration of EOATs. The standard covers end-effector design, suggested manufacturing principles, integration of an EOAT into a robot system, and necessary information for operation. Part 2 of ISO 20218 is dealing with manual load and unload stations that is out of scope for the red meat sector. 

The standard's main suggestion is to avoid dangerous structures by design on EOAT, e.g., sharp edges and pointed corners. Nonetheless, knives and saws are indispensable tools of meat-processing robot systems, thus the only option is risk-minimization. Risk reduction in such cases is mostly solved by physical protective devices and built-in safety-related control systems. Commonly used examples for the latter include capabilities for force sensing, speed monitoring, presence sensing, emergency stop. Besides covering sharp and pointed EOAT, ISO 20218 has a dedicated section for grippers, highlighting grasp-type grippers (force closure and form closure types), magnetic grippers and vacuum grippers. Grasp-type and vacuum grippers are commonly used in the meat-industry too, the RoBUTCHER project employs smart grasp-type grippers as EOAT and arrays of vacuum grippers as part of the robot system as well~\cite{grippersoa}. 

ISO 20218 contains annexes as well, including references to potentially relevant other standards and presenting real risk assessment scenarios. Furthermore, there are suggestions with examples for safety-rated monitored stopping, gripper safety performance assessment and a table about potential hazards, their possible origins and consequences.

\section{Food-safety standards}

In food industry automation projects, hygiene aspects and general food-safety are almost as important concerns as the safety of operators and machinery. Despite the fact that this article focuses on the safety regulations of food-sector robotics, food-safety should be mentioned as well, since the two topics are obviously strongly related~\cite{sodring2022effects}. ISO 22000:2018 Food safety management defines food safety as~\cite{ISO22000}:
\begin{quote}
\textit{Assurance that food will not cause an adverse health effect for the consumer when it is prepared and/or consumed in accordance with its intended use.}
\end{quote}
This comprehensive standard suggests the adoption of a Food Safety Management System (FSMS), claiming that it has a great potential in helping to develop a system’s performance regarding food safety~\cite{panghal2018}. The most important benefits of introducing an FSMS are:
\begin{itemize}
    \item The organization improves its ability to consistently provide safe products and food-related services that meet customer needs and satisfy all regulatory and statutory requirements;
    \item Identifying and addressing hazards associated with its products and services;
    \item The ability to demonstrate conformity to specified FSMS requirements.
\end{itemize}

\noindent The most general ISO standard in this topic is the ISO 22000:2018 Food safety management systems – requirements for any organization in the food chain~\cite{ISO22000}. The standard introduces a practical plan-do-check-act (PDCA) cycle in details, that shall be used at the development process of an FSMS. The PDCA should also be able to help with improving the FSMS's efficiency to achieve safe production and services along with fulfilling relevant requirements. 

The technical segment of ISO 22000 specifies the implementation of the PDCA cycle, offers suggestions about the tasks of the organization, and clarifies the communication, operation and documentation that is required to preserve the safety operation. The standard also covers hazard control, analysis and assessment, emergency response, monitoring and measuring. Its last sections offer possibilities and methods for internal auditing, review of management systems and continuous long-term improvement.

As ISO 22000 is a comprehensive standard, it suggests potentially relevant more specific standards, as well as other important official documents, such as: 

\begin{itemize}
\item ISO/TS 22002 Prerequisite programmes on food safety;
\item ISO/TS 22003 Food safety management systems — Requirements for bodies providing audit and certification of food safety management systems;
\item CAC/RCP 1-1969 General Principles of Food Hygiene;
\end{itemize}

\section{Discussion}
Although the number of machinery safety related standards for industrial robotics applications became rather large in the past decades, the selection for the fast growing domain of service robots is significantly smaller. Furthermore, there is no technically comprehensive standard for agri-food robotics applications, that would cover all safety aspects, and no specific standard that would cover meat industry automation. Regarding the automated meat processing applications with industrial robots (e.g., the Meat Factory Cell developed within the RoBUTCHER Project, see Fig.~\ref{fig:MFC_RS}), the general suggestion of the ISO standards is to follow the minimum hazard principle by methodically identifying and eliminating (or at least reducing) the risk factors. 

The most relevant ISO standard that was identified in this review is the ISO~10218, which is more than 10 years old, yet a new version is currently under development. According to this type-C standard, a systematic solution for safe design based on the existing relevant ISO standards should be possible to be given, even to novel, innovative robotic systems and applications. However, the Notified Body chosen to certify a new system might propose different or additional requirements. The adoption of already existing safety related guidelines from other domains e.g., from medical robotics, where safety requirements has been linked to the level of autonomy of a robotic system, to the food sector is currently the best-practise, and could be a beneficial method ~\cite{haidegger2019, chinzei2019}. Generally, the maximum safety control principle of a robot cell (e.g., development of advanced teleoperated systems instead of collaborative operations, especially when the robot cell contains remarkably dangerous tools) most likely will increase the applicability and deployability of such developments in the future. 

It is also worth mentioning that despite the general public (and official bodies as well) increasing support for sustainability in development of robotic applications with regulations~\cite{mai2022role,boesl2021automating}, the appropriate guidelines for streamlined implementations are still missing~\cite{mason2021meat}. Similarly, in spite of robot ethics becoming a general discussion topic, the establishment of proper standards and guidelines was just launched in the robotics and automation domains~\cite{prestes2021first,khamis2019ai}.

\section{Conclusion}

Nowadays, it is clear that automation and robotization mean the long-term solutions for many services and industrial applications. However, due to the complexity of the tasks in the food industry (agriculture and meat-processing as well), in many cases, it is still necessary to have human operators in the workplace too. This new kind of collaboration between humans and autonomous robots has elevated need for new and adapting safety features, thus for associated safety guidelines and standards too. The development of such regulations are in their early stages yet, however -- derived from the existing standards -- implementation of safety features will remain the manufacturer's responsibility, as s\textit{safety by design} is still the preferred design principle.

\subsection*{Acknowledgment}
This work has received funding from the European Union’s Horizon 2020 research and innovation programme under grant agreement No 871631, RoBUTCHER (A Robust, Flexible and Scalable Cognitive Robotics Platform). \\ P. Galambos's work is partially supported by Project no.~2019-1.3.1-KK-2019-00007, provided by the National Research, Development and Innovation Fund of Hungary. \\  T. Haidegger is a Bolyai Fellow of the Hungarian Academy of Sciences.\\ We acknowledge the assistance of RobotNorge AS with this topic as a partner in the RoBUTCHER project.

\subsection*{Abbreviations}{
The following abbreviations are used in this article:\\

\noindent 
\begin{tabular}{@{}ll}
CE & Conformité Européenne\\
CEO & Chief Executive Officer\\
DIH &  Digital Innovation Hub\\
DoF & Degrees of Freedom\\
EC MD & European Commission Machinery Directive\\
EOAT & End of Arm Tooling\\
EU & European Union\\
FSMS & Food Safety Management System\\
IEC & International Electrotechnical Commission\\
IMS & Integrated Manufacturing System\\
ISO & International Organization for Standardization\\
MAR & Multi-Annual Roadmap\\
PDCA & Plan-Do-Check-Act\\
RGB-D camera & Red-Green-Blue-Depth camera\\
TC & Technical Committee\\
TR & Technical Report

\end{tabular}
}

\newpage
\bibliography{actap_bib.bib}

\end{document}